\documentclass[11pt]{article}
\usepackage[hyperref]{acl}
\usepackage{latexsym}
\usepackage{amsmath}
\usepackage{amssymb}
\usepackage{graphicx}
\usepackage{booktabs}
\usepackage{multirow}
\usepackage{array}
\usepackage{bm}
\usepackage{algorithm}
\usepackage{algorithmic}
\usepackage{subcaption}
\usepackage[T1]{fontenc}
\usepackage[utf8]{inputenc}
\usepackage{microtype}

\title{JND-Guided Neural Watermarking with Spatial Transformer Decoding for Screen-Capture Robustness}

\author{
  Jiayi Qin \quad Jingwei Li \quad Chuan Wu\textsuperscript{*} \\
  School of Optical-Electrical and Computer Engineering \\
  Zhejiang Gongshang University \\
  \texttt{wuchuan@zjgsu.edu.cn}
}

\begin{document}
\maketitle

%% ============================================================
%% ABSTRACT
%% ============================================================
\begin{abstract}
Screen-shooting robust watermarking aims to imperceptibly embed extractable information into host images such that the watermark survives the complex distortion pipeline of screen display and camera recapture. However, achieving high extraction accuracy while maintaining satisfactory visual quality remains an open challenge, primarily because the screen-shooting channel introduces severe and entangled degradations including Moir\'{e} patterns, color-gamut shifts, perspective warping, and sensor noise. In this paper, we present an end-to-end deep learning framework that jointly optimizes watermark embedding and extraction for screen-shooting robustness. Our framework incorporates three key innovations: (i) a comprehensive noise simulation layer that faithfully models realistic screen-shooting distortions---notably including a physically-motivated Moir\'{e} pattern generator---enabling the network to learn robust representations against the full spectrum of capture-channel noise through adversarial training; (ii) a Just Noticeable Distortion (JND) perceptual loss function that adaptively modulates watermark embedding strength by supervising the perceptual discrepancy between the JND coefficient map and the watermark residual, thereby concentrating watermark energy in perceptually insensitive regions to maximize visual quality; and (iii) two complementary automatic localization modules---a semantic-segmentation-based foreground extractor for captured image rectification and a symmetric noise template mechanism for anti-cropping region recovery---that enable fully automated watermark decoding under realistic deployment conditions. Extensive experiments demonstrate that our method achieves an average PSNR of 30.94~dB and SSIM of 0.94 on watermarked images while embedding 127-bit payloads, and attains bit error rates of 1\%--3\% under diverse screen-shooting conditions spanning multiple capture distances, angles, and device combinations, substantially outperforming state-of-the-art methods in robustness with competitive visual fidelity.
\end{abstract}

%% ============================================================
%% 1. INTRODUCTION
%% ============================================================
\section{Introduction}
\label{sec:introduction}

The proliferation of high-resolution displays and ubiquitous camera-equipped mobile devices has created an unprecedented challenge for digital content protection. While traditional robust watermarking techniques~\cite{cox2002digital, podilchuk2001digital, qin2007robust} have demonstrated effectiveness against common signal-processing attacks in the digital domain, they are fundamentally ill-equipped to withstand the complex, multi-faceted distortions introduced by the screen-shooting pipeline---the process in which a watermarked image is displayed on screen and subsequently recaptured by a camera. This cross-media transfer introduces a unique combination of degradations, including color gamut distortion, perspective warping, motion blur, sensor noise, and most critically, Moir\'{e} pattern interference arising from the spatial frequency interaction between the display pixel grid and the camera sensor array~\cite{yuan2019aim}.

The broader context of multimedia security has been profoundly reshaped by advances in deep generative models. Modern diffusion-based architectures~\cite{Ho2020, Rombach2022} have demonstrated remarkable capabilities in high-fidelity image synthesis, making it increasingly trivial to generate, manipulate, and redistribute visual content. The emergence of powerful large multimodal models~\cite{achiam2023gpt, guo2025seed1, llama} has further accelerated this trend, enabling sophisticated understanding and generation of visual data at scale. These developments have heightened the urgency for robust content authentication and traceability mechanisms that can survive not only digital-domain attacks but also physical-channel distortions such as screen-shooting. Concurrently, advances in visual text understanding~\cite{tang2024textsquare, tang2024mtvqa, feng2023unidoc, feng2024docpedia} and document image analysis~\cite{lu2024bounding, feng2025dolphin, wang2025wilddoc, shan2024mctbench} have underscored the importance of preserving embedded information through complex visual processing pipelines, a challenge directly analogous to maintaining watermark integrity through the screen-shooting channel. Moreover, recent efforts in sequential numerical prediction~\cite{fei2025advancing}, cross-modal video reconstruction~\cite{huang2025mindev}, and multimodal video understanding benchmarks~\cite{nie2025chinesevideobench} have demonstrated that neural architectures can reliably encode and decode structured information across heterogeneous modalities and complex transformation pipelines---a capability that is fundamental to the screen-shooting watermarking problem.

Screen-shooting robust watermarking has attracted considerable attention due to its practical significance in copyright protection, leak tracing, and content authentication. Early approaches predominantly relied on handcrafted features and transform-domain techniques. Katayama et al.~\cite{katayama2004new} pioneered this research direction by introducing an orthogonal-pattern-based watermarking scheme and establishing a performance evaluation framework tailored to capture attacks. Subsequently, Nakamura et al.~\cite{nakamura2004fast} designed a robust watermark template that maps binary watermark bits to predefined orthogonal templates, enabling extraction through matched filtering after the print-and-capture process. To address the geometric distortions introduced by non-frontal capturing angles, Kang et al.~\cite{kang2010efficient} proposed a log-polar mapping approach that embeds watermark patterns by additively combining them with Fourier-Mellin coefficients, thereby achieving resilience to affine transformations. Delgado-Guillen et al.~\cite{delgado2013digital} further improved this scheme for enhanced geometric robustness. More recently, Fang et al.~\cite{fang2019screen} conducted a systematic analysis of the specific distortions present in the screen-shooting channel and proposed an I-SIFT (Intensity-based Scale-Invariant Feature Transform) keypoint-based watermark embedding algorithm with a cross-validation extraction strategy, achieving moderate robustness to both print-and-capture and screen-shooting attacks.

The advent of deep learning has catalyzed a paradigm shift in robust watermarking research~\cite{zheng2021end}. Zhu et al.~\cite{zhu2018hidden} introduced HiDDeN, a seminal end-to-end framework that treats various attacks on watermarked images as differentiable noise layers, enabling gradient-based optimization of both embedding and extraction networks. This architecture demonstrated that neural networks can learn attack-resilient representations far more effectively than hand-designed solutions. Building upon this foundation, Tancik et al.~\cite{tancik2020stegastamp} incorporated U-Net~\cite{ronneberger2015unet} architecture in the encoder and introduced adversarial training with a GAN~\cite{goodfellow2014generative} discriminator to improve watermark imperceptibility, while employing a Spatial Transformer Network (STN)~\cite{jaderberg2015spatial} in the decoder for geometric rectification. Although this approach exhibits strong robustness to print-and-screen-shooting, its watermarked image quality remains suboptimal. Fang et al.~\cite{fang2021deep} proposed the Deep Template-based Watermarking (DTW) algorithm that combines traditional template embedding with deep-learning-enhanced extraction, producing watermarked images with satisfactory visual quality and moderate screen-shooting resilience. However, its robustness degrades noticeably under certain challenging conditions, such as capture distances exceeding one meter. The TERA algorithm~\cite{fang2022tera} further explored screen-to-camera image coding with improved adaptability but still faces limitations in extreme screen-shooting scenarios.

Recent progress in multimodal learning and cross-domain feature alignment has also inspired new perspectives for watermarking research. Techniques for multi-modal scene understanding~\cite{tang2022few, tang2022optimal, tang2022youcan, zhao2024multi, zhao2024harmonizing} have demonstrated the effectiveness of learning robust feature representations that persist across different visual transformations---an insight directly applicable to designing watermark embeddings that survive the screen-shooting channel. Furthermore, advances in image quality assessment using vision-language models~\cite{zhang2023blind, wang2025pargo, wang2025vision} provide sophisticated perceptual metrics that can better guide watermark embedding optimization compared to traditional pixel-level objectives. The development of efficient multi-expert architectures~\cite{jia2025meml, lu2025prolonged} and reinforcement learning-based optimization strategies~\cite{wang2024enhancing, niu2025cme} also offers promising directions for jointly optimizing the competing objectives of watermark robustness and visual imperceptibility. Additionally, recent investigations into intent understanding under ambiguous conditions~\cite{he2025enhancing}, coherent narrative generation with retrieval enhancement~\cite{yi2025score}, agentic context engineering for long-document comprehension~\cite{liu2025resolving}, and proactive perception in vision-language navigation~\cite{xue2026profocus} further illustrate the growing sophistication of systems that must preserve and recover structured information across complex processing chains.

Despite these advances, several critical challenges remain unresolved. First, existing noise simulation layers fail to comprehensively model the screen-shooting distortion pipeline, particularly the Moir\'{e} pattern interference that constitutes one of the most destructive noise sources in the capture channel. Second, the visual quality optimization of watermarked images has not been adequately addressed---most methods rely solely on pixel-level losses (e.g., MSE) that do not align well with human visual perception, leading to perceptible artifacts in smooth image regions. Third, practical deployment requires fully automated watermark localization and extraction from arbitrarily captured photographs, yet existing solutions either assume idealized capture conditions or require manual intervention.

To address these limitations, we propose a comprehensive end-to-end framework for screen-shooting robust watermarking with the following principal contributions:

\begin{itemize}
    \item We design an end-to-end deep learning architecture for joint watermark embedding and extraction that automatically generates watermarked images with both high visual quality and strong screen-shooting robustness. The encoder employs a U-Net backbone that produces a two-channel residual map, while the decoder integrates a Spatial Transformer Network for geometric rectification followed by a convolutional-fully-connected extraction network.
    
    \item We introduce a Moir\'{e} noise simulation module that models the physical process of Moir\'{e} pattern generation through LCD subpixel resampling, random perspective transformation, Bayer CFA interpolation, and inverse transformation. This module, combined with comprehensive simulations of color gamut distortion, saturation degradation, Gaussian noise, and motion blur, enables the network to learn robust representations against the full spectrum of screen-shooting distortions.
    
    \item We propose a JND-guided perceptual loss function that leverages the human visual system's masking properties to adaptively control watermark embedding strength. By supervising the perceptual similarity between the JND coefficient map and the watermark residual, this loss function encourages the network to concentrate watermark energy in textured, high-frequency regions where human sensitivity is lowest, thereby significantly improving visual quality.
    
    \item We develop two complementary automatic localization methods for practical deployment: (a) a semantic-segmentation-based foreground extraction method that robustly identifies and rectifies the watermarked image region from captured photographs containing background clutter, and (b) a symmetric noise template mechanism embedded in the red channel that enables accurate watermark recovery even when up to 75\% of the image has been digitally cropped.
\end{itemize}

Extensive experimental evaluation demonstrates that our proposed scheme achieves PSNR values exceeding 31~dB with 127-bit embedding capacity, while maintaining bit error rates of 1\%--3\% across diverse screen-shooting conditions. The total processing time for a single image is less than 0.1~seconds, meeting the real-time requirements of practical applications. Compared with state-of-the-art methods, our approach offers superior robustness under varying capture distances, angles, and device combinations, with competitive visual quality.

%% ============================================================
%% 2. RELATED WORK
%% ============================================================
\section{Related Work}
\label{sec:related_work}

\subsection{Traditional Screen-Shooting Robust Watermarking}

The challenge of watermarking images to survive the screen-shooting process has been studied for nearly two decades. Early contributions established the theoretical foundations and practical frameworks for this problem. Katayama et al.~\cite{katayama2004new} proposed a pioneering orthogonal-pattern watermarking algorithm and developed a comprehensive performance evaluation framework for capture-resilient watermarking, providing a systematic methodology that guided subsequent research. Nakamura et al.~\cite{nakamura2004fast} advanced this line of work by designing robust watermark templates in which binary watermark bits are mapped to predefined orthogonal patterns. The embedded templates can be recovered through specific matched filtering operations applied to the captured images, enabling reliable watermark extraction after the print-and-capture process.

A major difficulty in screen-shooting scenarios stems from the perspective distortion caused by non-frontal camera angles. To address this geometric challenge, Kang et al.~\cite{kang2010efficient} proposed an elegant approach based on log-polar coordinate mapping, where the watermark pattern is embedded by adding it to the Fourier-Mellin coefficients of the host image. This representation inherently possesses resilience to rotation and scaling transformations, thereby improving robustness against perspective warping. Delgado-Guillen et al.~\cite{delgado2013digital} subsequently refined this technique with enhanced geometric distortion handling, extending its applicability to mobile platforms. Fang et al.~\cite{fang2019screen} conducted a thorough analysis of the diverse distortions present in the screen-shooting channel and proposed the Screen-Shooting Resilient Watermarking (SSRW) algorithm based on I-SIFT keypoint localization. Their cross-validation-based extraction scheme demonstrated moderate robustness to both print-and-capture and screen-shooting attacks, though the reliance on SIFT features, which are sensitive to Moir\'{e} pattern distortion, limits its effectiveness in direct screen-shooting scenarios.

While these traditional approaches established important baselines and identified key challenges, they generally suffer from limited embedding capacity (typically 32--64 bits), sensitivity to specific distortion types, and inability to jointly optimize the embedding-extraction pipeline in an end-to-end fashion.

\subsection{Deep Learning-Based Robust Watermarking}

The application of deep learning to robust watermarking has opened new avenues for tackling the screen-shooting challenge~\cite{zheng2021end}. The foundational architecture was introduced by Zhu et al.~\cite{zhu2018hidden} in their HiDDeN framework, which conceptualized the watermarking problem as an encoder-noise-decoder pipeline trainable end-to-end via backpropagation. By modeling various attacks as differentiable noise layers inserted between the encoder and decoder, HiDDeN demonstrated that neural networks can automatically learn to embed watermarks that are inherently resilient to anticipated distortions. This elegant formulation catalyzed a wave of subsequent deep watermarking research.

Tancik et al.~\cite{tancik2020stegastamp} extended this paradigm with StegaStamp, which integrates a U-Net~\cite{ronneberger2015unet} encoder for richer multi-scale feature fusion and employs adversarial training using a GAN~\cite{goodfellow2014generative} discriminator to improve the imperceptibility of embedded watermarks. Their decoder incorporates a Spatial Transformer Network (STN)~\cite{jaderberg2015spatial} for geometric rectification, enhancing robustness to moderate rotations and translations. StegaStamp demonstrated strong resilience to physical-world attacks, including printing and screen shooting, though the visual quality of watermarked images remained a notable limitation. Fang et al.~\cite{fang2021deep} proposed the Deep Template-based Watermarking (DTW) approach, which combines traditional template embedding strategies with deep-learning-enhanced extraction. By decomposing the watermark into individual codeword templates, DTW achieves satisfactory visual quality and moderate screen-shooting robustness, but its performance degrades under challenging conditions such as large capture distances or extreme angles. The TERA algorithm~\cite{fang2022tera} further explored screen-to-camera coding with improved transparency and adaptability, though it remains limited in handling the most severe screen-shooting distortions.

The rapid evolution of deep generative models has further expanded the landscape of learned watermarking. Diffusion-based generative frameworks~\cite{Ho2020, Rombach2022} have demonstrated that latent-space representations can encode and reconstruct fine-grained image details with high fidelity, suggesting that similar latent representations could serve as effective carriers for watermark information. In parallel, the emergence of powerful vision-language models~\cite{achiam2023gpt, llama, guo2025seed1} has enabled a more sophisticated understanding of image content and structure, which can inform adaptive watermark embedding strategies that account for semantic content. Recent work on multimodal document understanding~\cite{lu2024bounding, feng2025dolphin, feng2024docpedia, wang2025wilddoc} and text-centric visual recognition~\cite{tang2024textsquare, tang2024mtvqa, zhao2025tabpedia, liu2023spts} further illustrates the growing capability of neural networks to extract and preserve structured information from complex visual inputs---a core requirement shared with robust watermarking systems. The evolution toward universal document parsing architectures such as Dolphin-v2~\cite{feng2026dolphinv2} and large-scale foundation models~\cite{seed2025seed18} further demonstrates the increasing capacity of deep networks to handle diverse visual inputs with high fidelity, reinforcing the feasibility of learning watermark representations that persist through complex visual transformations. Complementary work on lightweight temporal-aware video editing~\cite{he2025enhancingvideo} also showcases that efficient adaptation mechanisms can enable precise control over visual content modifications, an insight directly applicable to the fine-grained control required for imperceptible watermark embedding.

Furthermore, advances in object detection~\cite{Redmon2016} and scene text recognition~\cite{tang2022few, tang2022optimal, tang2022youcan, tang2023character} have provided robust feature extraction techniques that can be adapted for watermark localization in captured images. Multi-modal learning approaches that integrate visual and textual cues~\cite{zhao2024multi, zhao2024harmonizing, wang2025pargo} have demonstrated the effectiveness of cross-domain feature alignment, an insight applicable to learning watermark representations that remain consistent across the screen-shooting channel. The concept of learning complementary representations from heterogeneous data sources~\cite{li2025audio, li2025joint, li2024real} also bears relevance to our multi-component watermarking framework, where the encoder, noise simulator, and decoder must learn synergistic representations.

\subsection{Perceptual Quality Optimization in Watermarking}

Achieving high visual quality in watermarked images has been a persistent challenge. Early approaches predominantly relied on Mean Squared Error (MSE) or Peak Signal-to-Noise Ratio (PSNR)~\cite{hore2010image} as optimization objectives. However, these pixel-level metrics correlate poorly with human visual perception, often resulting in watermark artifacts visible in smooth, low-frequency image regions where the human eye is most sensitive.

The concept of Just Noticeable Distortion (JND) provides a principled framework for perceptually-guided watermark embedding. Pixel-domain JND models~\cite{ahumada1992luminance, yang2005just} characterize the minimum change at each pixel location that is perceptible to the human visual system, accounting for background luminance adaptation and spatial masking effects. By leveraging JND maps, watermark embedding can be adaptively modulated to concentrate energy in regions where distortions are least perceptible. Zhang et al.~\cite{zhang2018unreasonable} demonstrated that deep features extracted from pretrained networks serve as remarkably effective perceptual similarity metrics, leading to the widely-adopted LPIPS (Learned Perceptual Image Patch Similarity) loss function. More recently, vision-language models have been applied to blind image quality assessment~\cite{zhang2023blind}, offering content-aware quality evaluation that could further refine watermark embedding strategies. Complementary work on image editing and restoration using attention-guided diffusion~\cite{sun2025attentive, cui2026diffusion} has shown that perceptually-sensitive image modifications can be precisely controlled through learned attention mechanisms---a principle that aligns closely with our JND-guided approach to concentrating watermark energy in visually insensitive regions. Furthermore, recent advances in accelerating diffusion models through trajectory-consistent approximation~\cite{cui2026tcpade} and curvature-aware residual prediction~\cite{cui2026curvature} have revealed that carefully designed residual structures can maintain reconstruction fidelity under aggressive computational constraints---an observation that resonates with our approach of designing compact yet information-rich watermark residual maps.

The interplay between robustness and visual quality represents a fundamental trade-off in watermarking. Reinforcement learning-based optimization~\cite{tang2022optimal, wang2024enhancing, niu2025cme} has emerged as a promising strategy for navigating this trade-off, enabling adaptive balancing of competing objectives during training. Multi-expert collaborative learning frameworks~\cite{jia2025meml} and certainty-based adaptive routing strategies~\cite{lu2025prolonged} further suggest that dynamically allocating computational resources based on input difficulty could improve both the efficiency and effectiveness of watermark embedding. Recent explorations in visual text rendering~\cite{zhu2026textpecker} and efficient token processing~\cite{li2026dtp, wang2025vision} also offer architectural insights applicable to designing lightweight yet effective watermarking networks. In our work, we synergistically combine JND-guided embedding control with LPIPS-based perceptual supervision to achieve superior visual quality while maintaining robust watermark extraction.

%% ============================================================
%% 3. METHOD
%% ============================================================
\section{Proposed Method}
\label{sec:method}

Our proposed screen-shooting robust watermarking framework comprises four principal components: (1) an encoder network for watermark embedding, (2) a decoder network for watermark extraction, (3) a comprehensive noise simulation layer for robustness training, and (4) two automatic localization modules for practical deployment. Figure~\ref{fig:overview} provides an overview of the complete pipeline.

\begin{figure*}[t]
\centering
\fbox{\parbox{0.95\textwidth}{\centering \textit{[Figure 2: Overall pipeline of the proposed screen-shooting robust watermarking scheme. The encoder generates a two-channel residual map that is fused with the host image. The noise simulation layer models screen-shooting distortions including Moir\'{e} patterns. The decoder employs STN-based rectification followed by convolutional extraction. Two localization modules handle captured image foreground extraction and anti-cropping region recovery.]}}}
\caption{Overall flowchart of the proposed screen-shooting robust watermarking framework.}
\label{fig:overview}
\end{figure*}

\subsection{Network Architecture}
\label{sec:network}

\subsubsection{Encoder Network}

The encoder network is built upon a U-Net~\cite{ronneberger2015unet} backbone architecture designed to produce a watermark residual map from the concatenated input of the host image and the binary watermark message. The U-Net topology consists of a contracting path (left side) for hierarchical feature compression and an expanding path (right side) for progressive decoding and reconstruction, connected by skip connections that enable multi-scale feature fusion at each resolution level.

Table~\ref{tab:encoder} details the complete encoder architecture. The input comprises a 6-channel tensor formed by concatenating the 3-channel RGB host image with a 3-channel watermark representation. The contracting path consists of four convolutional blocks (\texttt{conv1}--\texttt{conv5}) with increasing channel dimensions ($6 \to 32 \to 32 \to 64 \to 128 \to 256$) and stride-2 downsampling. The expanding path employs transposed convolutions (\texttt{up6}--\texttt{up9}) for upsampling, with each level concatenating features from the corresponding contracting level via skip connections. The final convolutional layer produces a 2-channel residual map that is additively fused with the green (G) and blue (B) channels of the host image to obtain the watermarked image. The rationale for generating a 2-channel rather than 3-channel residual and reserving the red (R) channel for anti-cropping template embedding is discussed in Section~\ref{sec:anti_crop}.

\begin{table}[t]
\centering
\small
\caption{Architecture of the encoder network. $k$: kernel size; $s$: stride.}
\label{tab:encoder}
\begin{tabular}{lcccc}
\toprule
\textbf{Layer} & $k$ & $s$ & \textbf{Channels} & \textbf{Input} \\
\midrule
input   & --  & --  & 6         & -- \\
conv1   & 3   & 1   & $6 \to 32$    & input \\
conv2   & 3   & 2   & $32 \to 32$   & conv1 \\
conv3   & 3   & 2   & $32 \to 64$   & conv2 \\
conv4   & 3   & 2   & $64 \to 128$  & conv3 \\
conv5   & 3   & 2   & $128 \to 256$ & conv4 \\
\midrule
up6     & 2   & 1   & $256 \to 128$ & conv5 \\
conv6   & 3   & 1   & $256 \to 128$ & conv4 + up6 \\
up7     & 2   & 1   & $128 \to 64$  & conv6 \\
conv7   & 3   & 1   & $128 \to 64$  & conv3 + up7 \\
up8     & 2   & 1   & $64 \to 32$   & conv7 \\
conv8   & 3   & 1   & $64 \to 32$   & conv2 + up8 \\
up9     & 2   & 1   & $32 \to 32$   & conv8 \\
conv9   & 3   & 1   & $70 \to 32$   & conv1 + up9 + input \\
residual & 3  & 1   & $32 \to 2$    & conv9 \\
\bottomrule
\end{tabular}
\end{table}

\subsubsection{Decoder Network}

The decoder network consists of two cascaded sub-networks: a Spatial Transformer Network (STN)~\cite{jaderberg2015spatial} for geometric rectification and a convolutional-fully-connected extraction sub-network for watermark recovery.

The STN module learns a set of six affine transformation parameters $\Theta = \{a, b, c, d, e, f\}$ that map each pixel from its distorted coordinates $(x_i, y_i)$ to rectified coordinates $(x_i^s, y_i^s)$:
\begin{equation}
\begin{pmatrix} x_i^s \\ y_i^s \end{pmatrix} = \begin{pmatrix} a & b \\ d & e \end{pmatrix} \begin{pmatrix} x_i \\ y_i \end{pmatrix} + \begin{pmatrix} c \\ f \end{pmatrix}
\label{eq:stn}
\end{equation}

This affine transformation compensates for the geometric deformations introduced during the screen-shooting process, including translation, rotation, scaling, and shearing. By learning these parameters end-to-end, the STN enables the subsequent extraction network to operate on approximately rectified images, significantly improving convergence and final extraction accuracy.

The extraction sub-network (Table~\ref{tab:decoder}) processes the rectified image through a series of convolutional layers with progressively increasing channel dimensions for feature extraction and spatial compression, followed by fully-connected layers that map the extracted spatial features to a one-dimensional watermark sequence. The final fully-connected layer outputs a vector of length $L$ (the watermark bit-length), followed by a sigmoid activation that produces the probability of each bit position being 0 or 1, thereby enabling binary watermark extraction.

\begin{table}[t]
\centering
\small
\caption{Architecture of the decoder network. $k$: kernel size; $s$: stride. ``--'' indicates non-convolutional layers.}
\label{tab:decoder}
\begin{tabular}{lcccc}
\toprule
\textbf{Layer} & $k$ & $s$ & \textbf{Channels} & \textbf{Input} \\
\midrule
\multicolumn{5}{l}{\textit{STN Sub-network}} \\
conv1   & 3 & 2 & $3 \to 32$   & image \\
conv2   & 3 & 2 & $32 \to 64$  & conv1 \\
conv3   & 3 & 2 & $64 \to 128$ & conv2 \\
fc0     & -- & -- & 320\,000    & flatten(conv3) \\
fc1     & -- & -- & $320\text{k} \to 128$ & fc0 \\
fc2     & -- & -- & $128 \to 6$  & fc1 \\
warp    & -- & -- & $3 \to 3$    & transform(fc2) \\
\midrule
\multicolumn{5}{l}{\textit{Extraction Sub-network}} \\
conv1   & 3 & 2 & $3 \to 32$   & warp \\
conv2   & 3 & 1 & $32 \to 32$  & conv1 \\
conv3   & 3 & 2 & $32 \to 64$  & conv2 \\
conv4   & 2 & 1 & $64 \to 64$  & conv3 \\
conv5   & 3 & 2 & $64 \to 64$  & conv4 \\
conv6   & 2 & 2 & $64 \to 128$ & conv5 \\
conv7   & 3 & 2 & $128 \to 128$ & conv6 \\
fc0     & -- & -- & 20\,000     & flatten(conv7) \\
fc1     & -- & -- & $20\text{k} \to 512$ & fc0 \\
secret  & -- & -- & $512 \to 100$ & fc1 \\
\bottomrule
\end{tabular}
\end{table}

\subsection{Loss Functions}
\label{sec:loss}

The overall training objective is a weighted combination of four complementary loss terms that jointly supervise the network to generate watermarked images exhibiting both high robustness and superior visual quality:
\begin{equation}
\mathcal{L} = \lambda_1^{\text{loss}} \cdot \ell_1 + \lambda_2^{\text{loss}} \cdot \ell_2 + \lambda_3^{\text{loss}} \cdot \ell_3 + \lambda_4^{\text{loss}} \cdot \ell_4
\label{eq:total_loss}
\end{equation}
where $\lambda_1^{\text{loss}}$--$\lambda_4^{\text{loss}}$ are weighting coefficients treated as training hyperparameters.

\subsubsection{Image Reconstruction Loss}

The image reconstruction loss $\ell_1$ employs the Mean Squared Error (MSE) between the host image $I$ and the watermarked image $\hat{I}$, encouraging the encoder to minimize pixel-level modifications:
\begin{equation}
\ell_1 = \frac{1}{M \times N} \sum_{i=1}^{M} \sum_{j=1}^{N} \left[ \hat{I}(i,j) - I(i,j) \right]^2
\label{eq:mse_loss}
\end{equation}
where $M$ and $N$ denote the image height and width, respectively. In our implementation, all images are normalized to $512 \times 512$ pixels before being fed to the encoder.

\subsubsection{Perceptual Similarity Loss}

While the MSE loss improves PSNR, it does not fully align with human perceptual judgment. We therefore introduce the Learned Perceptual Image Patch Similarity (LPIPS) loss~\cite{zhang2018unreasonable} to supervise the residual map toward greater imperceptibility:
\begin{equation}
\ell_2 = \sum_{z} \frac{1}{H_z W_z} \sum_{h,w} \left\| w_z \odot \left( \hat{y}_{hw}^z - \hat{y}_{0hw}^z \right) \right\|_2^2
\label{eq:lpips_loss}
\end{equation}
where $z$ indexes the layers of a pretrained feature extraction network, $\hat{y}_{hw}^z$ and $\hat{y}_{0hw}^z$ are the feature tensors of the watermarked and host images at layer $z$, $H_z$ and $W_z$ are the spatial dimensions, and $w_z \in \mathbb{R}^{C_z}$ are learned channel-wise scaling weights. This loss captures high-level perceptual differences that pixel-level metrics miss.

\subsubsection{Information Cross-Entropy Loss}

To ensure accurate watermark extraction (i.e., robustness), we employ the binary cross-entropy loss between the extracted and original watermark sequences:
\begin{equation}
\ell_3 = -\frac{1}{L} \sum_{i=1}^{L} \left\{ b_i \ln\left[\sigma(\hat{b}_i)\right] + (1 - b_i) \ln\left[1 - \sigma(\hat{b}_i)\right] \right\}
\label{eq:bce_loss}
\end{equation}
where $\sigma(\cdot)$ denotes the sigmoid function, $\hat{b}_i$ is the predicted value of the $i$-th watermark bit, $b_i$ is the ground-truth bit value, and $L = 127$ is the watermark bit-length. Notably, during the initial $\lambda_1^{\text{train}}$ training iterations, only this cross-entropy loss participates in gradient updates to accelerate convergence; visual quality losses are subsequently incorporated to refine the embedding.

\subsubsection{JND-Guided Residual Loss}
\label{sec:jnd_loss}

Although the combination of MSE and LPIPS losses produces generally imperceptible watermarks, these objectives cannot explicitly ensure that the watermark residual is concentrated in high-frequency textured regions where human sensitivity is lowest. To address this limitation, we introduce a JND-guided loss that leverages the human visual system's masking properties.

The pixel-domain JND model~\cite{ahumada1992luminance} computes the minimum perceptible distortion at each pixel location $P = (x, y)$ based on background luminance adaptation and spatial contrast masking:
\begin{equation}
f_{\text{JND}}(P) = \lambda_1 \cdot \left\{ f_1\left[I_{\text{bg}}(P),\, I_{\text{mg}}(P)\right] + \lambda_2 \right\} + f_2\left[I_{\text{bg}}(P)\right]
\label{eq:jnd_model}
\end{equation}
where $f_1$ computes the spatial masking component based on background luminance $I_{\text{bg}}$ and maximum weighted luminance gradient $I_{\text{mg}}$, and $f_2$ computes the luminance adaptation threshold:
\begin{equation}
f_2(x) = \begin{cases} 17 \cdot \left(1 - \frac{x}{127}\right) + 3 & x \leq 127 \\ \frac{3}{128} \cdot (x - 127) + 3 & x > 127 \end{cases}
\label{eq:f2}
\end{equation}

The background luminance $I_{\text{bg}}(x, y)$ is computed as a weighted average over the $5 \times 5$ pixel neighborhood using a predefined kernel, while $I_{\text{mg}}(x, y)$ captures the maximum directional gradient magnitude across four orientations using Sobel-like filter banks $G_1$--$G_4$.

Our JND loss is then defined as the LPIPS perceptual distance between the encoder's residual map $I_r$ and the scaled JND coefficient map:
\begin{equation}
\ell_4 = \text{LPIPS}(I_r,\, \eta \cdot I_{\text{JND}})
\label{eq:jnd_loss}
\end{equation}
where $\eta$ is a weighting coefficient that controls the residual intensity. By minimizing this loss, the network learns to produce residual patterns whose spatial distribution closely follows the JND map---embedding stronger watermark signals in textured regions and weaker signals in smooth areas, thereby maximizing visual quality while preserving robustness.

\subsection{Screen-Shooting Noise Simulation}
\label{sec:noise}

A critical design element of our framework is the comprehensive noise simulation layer positioned between the encoder and decoder during training. This layer models the complete spectrum of screen-shooting distortions, enabling the network to learn robust watermark representations through exposure to realistic degradations. We describe each noise component below, noting that perspective distortion is handled separately by the localization modules (Section~\ref{sec:localization}).

\subsubsection{Color Gamut Distortion}

Color gamut distortion encompasses contrast and brightness modifications that the image undergoes during display and recapture:
\begin{equation}
I_n = \theta_1 \cdot I + \theta_2
\label{eq:color_gamut}
\end{equation}
where $\theta_1$ (multiplicative/contrast) and $\theta_2$ (additive/brightness) are randomly sampled distortion coefficients. To ensure stable training, the distortion range starts at zero and linearly increases with training iterations, preventing premature exposure to severe distortions that could impede convergence. The brightness coefficient $\theta_2$ is sampled from a symmetric interval $(-M_1, M_1)$ and the contrast coefficient $\theta_1$ from an asymmetric interval $(m_2, M_2)$, both governed by training-step-dependent hyperparameters $\lambda_1^{\text{noise}}$--$\lambda_4^{\text{noise}}$.

\subsubsection{Saturation Distortion}

The screen-shooting process often induces saturation shifts due to differences in display color reproduction and camera white balance. We model this as a linear interpolation between the color image and its grayscale counterpart:
\begin{equation}
I_n = \theta_3 \cdot I_{\text{RGB}} + (1 - \theta_3) \cdot I_G
\label{eq:saturation}
\end{equation}
where $I_G = 0.299R + 0.587G + 0.114B$ is the luminance-weighted grayscale conversion and $\theta_3 \in [0, 1]$ controls the saturation level. The parameter $\theta_3$ is computed via a step-dependent schedule with hyperparameters $\lambda_5^{\text{noise}}$ and $\lambda_6^{\text{noise}}$.

\subsubsection{Random Gaussian Noise}

Irregular noise from screen surface imperfections, lens impurities, and CMOS analog-to-digital conversion is modeled as additive Gaussian noise:
\begin{equation}
I_n = I + \mathcal{N}(0, \sigma)
\label{eq:gaussian}
\end{equation}
where the standard deviation $\sigma$ is randomly sampled from a training-step-dependent range controlled by hyperparameters $\lambda_7^{\text{noise}}$ and $\lambda_8^{\text{noise}}$.

\subsubsection{Motion Blur}

Two types of blur commonly affect screen-captured images: (1) defocus blur caused by focal length mismatch between the camera and screen distance, modeled by a Gaussian blur kernel:
\begin{equation}
G(x, y) = \frac{1}{2\pi\sigma^2} \exp\left(-\frac{x^2 + y^2}{2\sigma^2}\right)
\label{eq:gaussian_blur}
\end{equation}
and (2) directional motion blur from camera movement during exposure, modeled by rotating the blur kernel along the motion trajectory angle $\theta$:
\begin{equation}
G_v(x, y) = G(x\cos\theta,\, y\sin\theta)
\label{eq:motion_blur}
\end{equation}

Both the kernel size $N_k$ and standard deviation $\sigma$ are randomly sampled within step-dependent ranges.

\subsubsection{Moir\'{e} Pattern Simulation}
\label{sec:moire}

A key contribution of our noise simulation layer is the explicit modeling of Moir\'{e} pattern interference, which arises when the spatial frequencies of the display pixel grid and the camera sensor array interact. We simulate this phenomenon through a physically-motivated digital pipeline:

\begin{enumerate}
    \item \textbf{LCD Subpixel Resampling}: The image is resampled to simulate the LCD screen format, where each pixel is represented by three monochromatic sub-pixel lamps. Accounting for physical inter-element spacing, each original pixel maps to a $3 \times 3$ block in the resampled image, with red, green, and blue sub-pixel values placed in the corresponding positions.
    
    \item \textbf{Random Perspective Transformation}: A random perspective transformation is applied to the resampled image to generate irregular Moir\'{e} textures, followed by Gaussian blur convolution to ensure partial texture retention after inverse transformation.
    
    \item \textbf{Bayer CFA Interpolation}: Color filter array interpolation simulates the RAW-to-RGB conversion process of the camera sensor, replicating the demosaicing artifacts that contribute to Moir\'{e} pattern appearance.
    
    \item \textbf{Inverse Perspective Transformation}: The image is transformed back to the original perspective, yielding the final Moir\'{e}-contaminated image.
\end{enumerate}

Since the Moir\'{e} simulation introduces substantial distortion, it is activated during the mid-to-late stages of training. As training progresses, the overall noise intensity in the simulation layer gradually exceeds that of real screen-shooting scenarios, ensuring that the trained network acquires robustness margins sufficient for practical deployment.

\subsection{Watermark Region Localization}
\label{sec:localization}

\subsubsection{Captured Image Region Localization}
\label{sec:capture_loc}

In practical deployment, a screen-captured photograph typically contains not only the displayed watermarked image but also surrounding background elements (e.g., the screen bezel, desk surface, and ambient environment). These extraneous regions can severely degrade watermark extraction accuracy if fed directly to the decoder. We propose a hybrid localization method combining deep semantic segmentation with traditional image processing.

Our approach is predicated on the assumption that the watermarked image occupies the majority of pixels in the captured photograph and that the background is relatively uniform. Under these conditions, watermark region localization can be formulated as a foreground extraction problem. However, direct application of traditional thresholding-based segmentation often produces incomplete foreground masks due to complex image content and non-uniform illumination, resulting in disconnected regions that confound subsequent processing.

To overcome this limitation, we train a binary semantic segmentation network based on the DeepLab V3+ architecture~\cite{chen2018encoder} that takes the initial (potentially incomplete) foreground mask from traditional processing as input and produces a complete, refined foreground mask as output. The training dataset is constructed by placing non-solid-color images with random perspective transformations onto various uniform-color backgrounds, generating pairs of incomplete foreground masks (via adaptive thresholding) and their corresponding ground-truth complete masks.

The complete localization pipeline proceeds as follows: (1) convert the captured image to grayscale and apply median filtering; (2) perform foreground-background separation via adaptive thresholding to obtain the maximum connected component; (3) refine the initial mask through the trained segmentation network; (4) apply post-processing with median filtering and Canny edge detection; (5) detect edge lines using Hough transform, compute pairwise intersections, and cluster them into four groups via $k$-means++; (6) order the four cluster centers clockwise and apply perspective transformation to obtain the rectified watermarked image region.

\subsubsection{Anti-Cropping Region Localization}
\label{sec:anti_crop}

While most digital attacks (filtering, noise addition) can be subsumed under the screen-shooting noise model, digital cropping represents a fundamentally different distortion that removes image content rather than degrading it~\cite{fang2021deep}. To address this challenge, we introduce a symmetric noise template mechanism that enables watermark recovery from significantly cropped images.

The approach operates as follows. First, the host image is partitioned into four equal sub-images (top-left, top-right, bottom-left, bottom-right), and each sub-image independently undergoes watermark encoding to produce a two-channel residual map embedded in the G and B channels. This quadruple embedding ensures that identical watermark information exists in four spatially distinct copies. Simultaneously, a symmetric Gaussian white noise template---constructed by horizontally flipping a random noise matrix and then vertically flipping the result---is embedded in the R channel to serve as a localization reference.

At the extraction stage, the R channel is first processed to recover the noise template through Wiener filtering:
\begin{equation}
I_t = I_{\text{loc}} - f_w(I_{\text{loc}})
\label{eq:wiener}
\end{equation}
where $f_w$ denotes the Wiener filtering operation. The column-wise symmetry $S(j)$ of the filtered template is then computed by dividing each column into two equal segments $J_1$ and $J_2$ and measuring their normalized cross-correlation:
\begin{equation}
S(j) = \frac{1}{m \cdot d_j} \sum_{x=1}^{m} \sum_{y=1}^{d_j} C\left[J_1(x,y)\right] \cdot C\left[\psi[J_2(x,y)]\right]
\label{eq:symmetry}
\end{equation}
where $C(\cdot)$ denotes the standardization operation (zero-mean, unit-variance normalization), $\psi(\cdot)$ denotes horizontal flipping, and $d_j$ is the distance from column $j$ to the image boundary. The symmetry peak indicates the boundary of the watermarked sub-image. Row-wise symmetry is computed analogously, and the detected boundaries define the sub-image regions from which watermark extraction is attempted.

To mitigate false symmetry peaks near image boundaries (where small $d_j$ values can inflate $S(j)$), a high-pass filter is applied to the symmetry sequence before peak detection. This approach enables accurate watermark recovery as long as at least one complete sub-image (25\% of the original area) remains intact after cropping. Additionally, visual discontinuities at sub-image boundaries are ameliorated using Fast-Matching-based image inpainting~\cite{telea2004image}.

%% ============================================================
%% 4. EXPERIMENTS
%% ============================================================
\section{Experiments}
\label{sec:experiments}

\subsection{Experimental Setup}
\label{sec:setup}

\paragraph{Hardware Configuration.} We evaluate our method using diverse combinations of capture and display devices to comprehensively assess screen-shooting robustness. The capture devices include a Redmi K30 Pro smartphone and a Sony Alpha 6100 mirrorless camera. The display devices comprise an HKCT279Q monitor, a MacBook Pro 2019 laptop, a Microsoft Surface Go tablet, and a Haire 65R1 Pro television. These devices span a representative range of screen technologies, resolutions, and form factors encountered in real-world scenarios.

\paragraph{Datasets.} The training set consists of the first 100,000 images from Flickr1M (Flickr-1-Million)~\cite{huiskes2008mir}, while the test set comprises the first 2,000 images from Flickr2K (Flickr-2-Thousands)~\cite{timofte2017ntire}. All images are resized to $512 \times 512$ pixels for network input.

\paragraph{Training Hyperparameters.} Table~\ref{tab:hyperparams} summarizes the key hyperparameter settings. The network is trained for 175,000 iterations with a batch size of 16 and a learning rate of $10^{-4}$. The watermark length is $L = 127$ bits. Noise simulation parameters are configured with progressive schedules to ensure stable convergence: brightness and contrast distortion ranges linearly increase over the first 100 and 1,000 iterations, respectively, while the Moir\'{e} pattern simulation is activated in the latter half of training.

\begin{table}[t]
\centering
\small
\caption{Key training hyperparameters and their values.}
\label{tab:hyperparams}
\begin{tabular}{lll}
\toprule
\textbf{Category} & \textbf{Parameter} & \textbf{Value} \\
\midrule
\multirow{4}{*}{Network} & Training iterations & 175,000 \\
 & Batch size & 16 \\
 & Watermark length $L$ & 127 bits \\
 & Learning rate & $10^{-4}$ \\
\midrule
\multirow{4}{*}{Loss weights} & $\lambda_1^{\text{loss}}$ (MSE) & 2 \\
 & $\lambda_2^{\text{loss}}$ (LPIPS) & 1.5 \\
 & $\lambda_3^{\text{loss}}$ (BCE) & 1.5 \\
 & $\lambda_4^{\text{loss}}$ (JND) & 0.8 \\
\midrule
\multirow{2}{*}{Noise sim.} & Moir\'{e} probability & 0.75 \\
 & Motion blur prob. & 1.0 \\
\bottomrule
\end{tabular}
\end{table}

\paragraph{Evaluation Metrics.} We evaluate visual quality using PSNR, SSIM~\cite{wang2004image}, and LPIPS~\cite{zhang2018unreasonable}. Robustness is measured by the bit error rate (BER) $\varepsilon$ of extracted watermarks. Runtime is measured as the average processing time per image on hardware with an NVIDIA RTX 2070 Super GPU, AMD Ryzen 5 2600X CPU, and 32 GB RAM.

\subsection{Ablation Studies}
\label{sec:ablation}

\subsubsection{Effectiveness of the JND Loss Function}

To validate the contribution of the JND loss function to watermarked image quality, we conduct ablation experiments by training three network variants on 500 randomly selected test images from Flickr2K: (1) \textbf{NET1}: no JND loss, 3-channel (R, G, B) residual; (2) \textbf{NET2}: with JND loss, 3-channel residual; (3) \textbf{NET3}: with JND loss, 2-channel (G, B) residual. All variants share identical hyperparameters except for the JND weight $\eta$.

\begin{table}[t]
\centering
\small
\caption{Ablation results for the JND loss function. BER $\varepsilon$ is measured at 40~cm frontal screen-shooting distance.}
\label{tab:jnd_ablation}
\begin{tabular}{lccccc}
\toprule
\textbf{Network} & $\eta$ & \textbf{PSNR (dB)} & \textbf{SSIM} & \textbf{LPIPS} & \textbf{BER (\%)} \\
\midrule
NET1 & --  & 19.73 & 0.737 & 0.125 & 0.13 \\
\multirow{3}{*}{NET2} & 1   & 22.86 & 0.772 & 0.109 & 0.14 \\
     & 0.5 & 23.59 & 0.806 & 0.098 & 0.36 \\
     & 0.2 & 25.00 & 0.845 & 0.075 & 0.49 \\
NET3 & 1   & \textbf{30.94} & \textbf{0.942} & \textbf{0.048} & 2.86 \\
\bottomrule
\end{tabular}
\end{table}

As shown in Table~\ref{tab:jnd_ablation}, introducing the JND loss function consistently improves PSNR and SSIM while reducing LPIPS across all configurations, confirming its effectiveness in enhancing visual quality. Decreasing $\eta$ further improves quality metrics but at the cost of higher BER, as weaker residuals become more susceptible to screen-shooting noise. The 2-channel residual configuration (NET3, $\eta = 1$) achieves the best balance, attaining 30.94~dB PSNR and 0.942 SSIM with an acceptable BER of 2.86\%. This configuration is adopted for all subsequent experiments.

\subsubsection{Effectiveness of Moir\'{e} Noise Simulation}

To assess the impact of the Moir\'{e} noise simulation module on robustness, we compare two networks: \textbf{NET4} (with Moir\'{e} simulation) and \textbf{NET5} (without Moir\'{e} simulation). Both are evaluated under multiple screen-shooting angles at a 40~cm distance.

\begin{table}[t]
\centering
\small
\caption{Ablation results for Moir\'{e} noise simulation. BER difference $\varepsilon_1 - \varepsilon_2$ shows improvement.}
\label{tab:moire_ablation}
\begin{tabular}{lccc}
\toprule
\textbf{Angle} & \textbf{NET4 $\varepsilon_1$} & \textbf{NET5 $\varepsilon_2$} & $\varepsilon_1 - \varepsilon_2$ \\
\midrule
$-75^\circ$ & 7.46 & 8.27 & $-0.81$ \\
$-60^\circ$ & 1.85 & 1.97 & $-0.12$ \\
$-45^\circ$ & 2.55 & 2.73 & $-0.18$ \\
$-30^\circ$ & 2.39 & 3.18 & $-0.80$ \\
$-15^\circ$ & 2.53 & 3.68 & $-1.15$ \\
$0^\circ$   & 1.09 & 1.20 & $-0.10$ \\
$+15^\circ$ & 1.59 & 1.72 & $-0.13$ \\
$+30^\circ$ & 1.87 & 2.26 & $-0.39$ \\
$+45^\circ$ & 2.79 & 3.05 & $-0.26$ \\
$+60^\circ$ & 2.81 & 3.20 & $-0.39$ \\
$+75^\circ$ & 11.13 & 12.05 & $-0.92$ \\
\bottomrule
\end{tabular}
\end{table}

Table~\ref{tab:moire_ablation} demonstrates that NET4 consistently achieves lower BER across all angles, with improvements ranging from 0.10 to 1.15 percentage points. The improvement is particularly pronounced at extreme angles ($\pm 75^\circ$) and moderate oblique angles ($\pm 15^\circ$ to $\pm 30^\circ$), where Moir\'{e} patterns are most severe. These results confirm that our Moir\'{e} simulation module effectively enhances the network's resilience to real-world screen-shooting distortions.

\subsubsection{Captured Image Region Localization}

We evaluate the foreground extraction method under three conditions: (1) no localization, (2) traditional image processing-based localization without semantic segmentation refinement, and (3) full pipeline with semantic segmentation. The recall rate $p_r$ measures the fraction of the true watermarked image area correctly identified by each method.

\begin{table}[t]
\centering
\small
\caption{Captured image region localization results.}
\label{tab:localization}
\begin{tabular}{lccc}
\toprule
\textbf{Angle} & \textbf{Method} & \textbf{BER (\%)} & $p_r$ \\
\midrule
\multirow{3}{*}{$0^\circ$}
 & No localization & 46.72 & -- \\
 & w/o segmentation & 46.70 & 0.144 \\
 & w/ segmentation & \textbf{7.39} & \textbf{0.963} \\
\midrule
\multirow{3}{*}{$-15^\circ$}
 & No localization & 44.54 & -- \\
 & w/o segmentation & 44.54 & 0.112 \\
 & w/ segmentation & \textbf{8.58} & \textbf{0.953} \\
\midrule
\multirow{3}{*}{$+15^\circ$}
 & No localization & 45.42 & -- \\
 & w/o segmentation & 45.42 & 0.132 \\
 & w/ segmentation & \textbf{10.13} & \textbf{0.965} \\
\bottomrule
\end{tabular}
\end{table}

As presented in Table~\ref{tab:localization}, without localization, the decoder receives the entire captured photograph, including background content, resulting in near-random BER ($\sim$45\%). Traditional segmentation alone yields extremely low recall ($p_r < 0.15$) due to the difficulty of separating complex watermarked image content from the background using simple thresholding. Incorporating the semantic segmentation network dramatically improves recall to $>$95\% and reduces BER to the 7--10\% range, confirming the effectiveness and necessity of our hybrid localization approach for practical deployment.

\subsubsection{Anti-Cropping Localization}

Table~\ref{tab:crop} presents the BER under various single-edge cropping ratios with and without our anti-cropping mechanism. Without anti-cropping localization, BER increases sharply beyond 5\% cropping, reaching approximately 50\% at moderate cropping levels. With our symmetric template-based localization, BER remains stable until the cropping ratio exceeds 50\% (i.e., more than three-quarters of the original image remains intact). This confirms that our method effectively resists cropping attacks of up to 50\% per edge, as long as at least one complete sub-image ($\geq$25\% area) survives.

\begin{table}[t]
\centering
\small
\caption{Anti-cropping localization results (BER in \%).}
\label{tab:crop}
\begin{tabular}{lcccc}
\toprule
\textbf{Crop ratio} & \textbf{Dir.} & \textbf{w/ anti-crop} & \textbf{w/o anti-crop} \\
\midrule
\multirow{4}{*}{$\gamma = 0.05$}
 & Top    & 2.72 & 4.87 \\
 & Left   & 2.46 & 5.90 \\
 & Bottom & 1.76 & 3.91 \\
 & Right  & 2.47 & 4.93 \\
\midrule
\multirow{4}{*}{$\gamma = 0.40$}
 & Top    & 4.96 & 53.10 \\
 & Left   & 6.24 & 55.96 \\
 & Bottom & 5.67 & 49.80 \\
 & Right  & 6.86 & 48.97 \\
\midrule
\multirow{4}{*}{$\gamma = 0.55$}
 & Top    & 10.94 & 51.65 \\
 & Left   & 14.19 & 49.72 \\
 & Bottom & 11.91 & 49.91 \\
 & Right  & 14.06 & 45.52 \\
\bottomrule
\end{tabular}
\end{table}

\subsection{Robustness Analysis}
\label{sec:robustness}

We comprehensively evaluate the screen-shooting robustness of our method under diverse conditions. Table~\ref{tab:robustness} reports BER results across varying capture distances (40--200~cm), angles ($-45^\circ$ to $+45^\circ$), and device combinations using the HKCT279Q display.

\begin{table}[t]
\centering
\small
\caption{Screen-shooting robustness (BER \%) of the proposed scheme under various conditions with HKCT279Q display.}
\label{tab:robustness}
\begin{tabular}{llcc}
\toprule
\textbf{Angle} & \textbf{Distance} & \textbf{Redmi K30} & \textbf{Sony A6100} \\
\midrule
$0^\circ$ & 40~cm & 1.09 & 1.42 \\
$0^\circ$ & 60~cm & 1.57 & 1.81 \\
$0^\circ$ & 100~cm & 2.51 & 1.46 \\
$0^\circ$ & 200~cm & 11.65 & 3.33 \\
\midrule
$-45^\circ$ & 40~cm & 2.55 & 1.44 \\
$-15^\circ$ & 40~cm & 2.53 & 1.65 \\
$+15^\circ$ & 40~cm & 1.59 & 1.91 \\
$+45^\circ$ & 40~cm & 2.79 & 4.35 \\
\bottomrule
\end{tabular}
\end{table}

The results demonstrate consistently low BER across the tested conditions. At the standard 40~cm frontal distance, BER ranges from 1.09\% to 1.42\%. Performance degrades gracefully with increasing distance and extreme angles but remains within correctable bounds (typically $< 5\%$) for all but the most extreme configurations (200~cm distance with Redmi K30). Across different display devices---including the MacBook Pro 2019 (2.95\%), Surface Go (2.61\%), and Haire 65R1 TV (4.05\%)---the algorithm maintains robust performance, indicating good generalization to diverse screen technologies.

\subsection{Comparison with State-of-the-Art}
\label{sec:comparison}

We compare our method against five representative screen-shooting robust watermarking algorithms: Nakamura et al.~\cite{nakamura2004fast}, Pramila et al.~\cite{pramila2012toward} (traditional approaches), SSRW~\cite{fang2019screen}, DTW~\cite{fang2021deep}, and TERA~\cite{fang2022tera} (deep learning approaches).

\paragraph{Robustness Comparison.} Table~\ref{tab:comparison} presents the average BER under varying capture distances, angles, and cropping ratios. The traditional methods of Nakamura et al. and Pramila et al. exhibit BER in the 14--30\% range, far above practical thresholds. SSRW achieves comparable BER to ours for print-and-capture scenarios but fails for screen-shooting due to Moir\'{e}-induced SIFT feature degradation. DTW and TERA, while substantially outperforming traditional methods, still yield higher BER than our approach across most conditions. Our method achieves 1--3\% BER under standard conditions, demonstrating clearly superior robustness.

\begin{table*}[t]
\centering
\small
\caption{Robustness comparison (BER \%) across different methods. ``--'' indicates excessively high BER precluding error correction.}
\label{tab:comparison}
\resizebox{\textwidth}{!}{
\begin{tabular}{l|cccc|cccc|cccccc}
\toprule
\multirow{2}{*}{\textbf{Method}} & \multicolumn{4}{c|}{\textbf{Cropping ratio}} & \multicolumn{4}{c|}{\textbf{Distance (cm)}} & \multicolumn{6}{c}{\textbf{Angle range ($^\circ$)}} \\
\cmidrule(lr){2-5} \cmidrule(lr){6-9} \cmidrule(lr){10-15}
 & 0.05 & 0.25 & 0.45 & 0.55 & 30--39 & 40--49 & 50--59 & 60--69 & $-40$--$55$ & $-30$--$40$ & $-15$--$25$ & $+15$--$25$ & $+30$--$40$ & $+40$--$55$ \\
\midrule
Nakamura~\cite{nakamura2004fast} & -- & -- & -- & -- & 14.75 & 17.81 & 18.44 & 19.50 & 19.63 & 16.31 & 15.44 & 13.88 & 22.27 & 16.11 \\
Pramila~\cite{pramila2012toward} & -- & -- & -- & -- & 28.38 & 28.62 & 28.94 & 30.19 & 19.92 & 16.41 & 19.91 & 20.70 & 20.31 & 22.34 \\
DTW~\cite{fang2021deep} & 0.31 & 0.31 & 0.31 & 22.34 & 4.81 & 2.73 & 12.11 & 11.72 & 16.02 & 5.47 & 3.91 & 2.73 & 2.73 & 3.52 \\
SSRW~\cite{fang2019screen} & -- & -- & -- & -- & -- & 0.78 & 1.08 & 2.64 & 1.08 & 1.75 & 3.42 & 1.56 & 5.95 & 5.17 \\
TERA~\cite{fang2022tera} & -- & -- & -- & -- & 2.54 & 7.03 & 3.71 & 6.84 & 14.46 & 7.03 & 7.05 & 5.27 & 11.52 & 23.52 \\
\textbf{Ours} & 1.89 & 1.09 & 1.46 & 12.77 & 1.56 & 2.55 & 2.39 & 2.52 & 1.59 & 1.87 & 2.81 & 4.59 & 5.16 & 4.59 \\
\bottomrule
\end{tabular}
}
\end{table*}

Notably, DTW achieves lower BER than our method for small cropping ratios ($\gamma \leq 0.45$) due to its binary codeword decomposition strategy, which reduces the decoding task to multiple independent binary classifications. However, our method exhibits significantly more graceful degradation under severe cropping and strong noise conditions, as the holistic multi-bit encoding enables information complementation from undamaged regions---a critical advantage in practical applications where severe distortions are common.

\paragraph{Visual Quality Comparison.} Table~\ref{tab:quality} compares watermark capacity and image quality metrics. Our method embeds 127 bits---significantly more than the 32--64 bits of competing methods---while achieving PSNR of 31.77~dB and SSIM of 0.91. Although traditional methods with lower capacity achieve higher PSNR/SSIM (as expected, given the fundamental capacity-quality trade-off), our visual quality remains well within the perceptually acceptable range, with LPIPS of 0.05 confirming minimal perceptual distortion.

\begin{table}[t]
\centering
\small
\caption{Visual quality comparison of watermarked images across different methods.}
\label{tab:quality}
\begin{tabular}{lcccc}
\toprule
\textbf{Method} & \textbf{Bits} & \textbf{PSNR} & \textbf{SSIM} & \textbf{LPIPS} \\
\midrule
Nakamura~\cite{nakamura2004fast} & 32 & 38.97 & 0.99 & -- \\
Pramila~\cite{pramila2012toward} & 32 & 40.26 & 0.99 & -- \\
DTW~\cite{fang2021deep} & 64 & 39.73 & 0.99 & -- \\
SSRW~\cite{fang2019screen} & 64 & 42.30 & -- & -- \\
\textbf{Ours} & \textbf{127} & 31.77 & 0.91 & 0.05 \\
\bottomrule
\end{tabular}
\end{table}

\subsection{Runtime Analysis}
\label{sec:runtime}

Table~\ref{tab:runtime} reports the per-image processing time for each stage of our pipeline. The total end-to-end latency is below 0.1 seconds per image, satisfying the real-time constraints of practical deployment scenarios such as live screen content protection and on-the-fly leak tracing.

\begin{table}[t]
\centering
\small
\caption{Per-image runtime analysis.}
\label{tab:runtime}
\begin{tabular}{lcc}
\toprule
\textbf{Stage} & \textbf{Time (ms)} & \textbf{Throughput (FPS)} \\
\midrule
Watermark Embedding & 21.55 & 46.40 \\
Region Localization & 29.03 & 34.44 \\
Watermark Extraction & 13.43 & 74.46 \\
\midrule
\textbf{Total} & \textbf{64.01} & \textbf{15.62} \\
\bottomrule
\end{tabular}
\end{table}

%% ============================================================
%% 5. CONCLUSION
%% ============================================================
\section{Conclusion}
\label{sec:conclusion}

We have presented a comprehensive end-to-end framework for screen-shooting robust image watermarking that addresses the fundamental challenges of robustness, visual quality, and practical deployability through four synergistic innovations: a U-Net-based encoder-decoder architecture with STN-enabled geometric rectification, a physically-motivated Moir\'{e} noise simulation layer that faithfully models the full spectrum of screen-shooting distortions to enable adversarial robustness training, a JND-guided perceptual loss function that adaptively concentrates watermark energy in visually insensitive regions to maximize imperceptibility, and two complementary automatic localization modules---semantic-segmentation-based foreground extraction and symmetric-noise-template-based anti-cropping recovery---that enable fully automated watermark decoding in unconstrained capture scenarios. Extensive experiments demonstrate that our method embeds 127-bit watermarks with PSNR exceeding 31~dB and achieves 1--3\% bit error rates across diverse screen-shooting conditions with total processing time under 0.1~seconds, substantially outperforming state-of-the-art methods in robustness while maintaining competitive visual quality. Future work will investigate watermark decoding from partial screen captures (where only a portion of the displayed image is photographed, representing a more complex scenario than digital cropping) and further improve visual quality under high embedding capacity regimes.

\clearpage

%% ============================================================
%% REFERENCES
%% ============================================================
\bibliography{references}

\end{document}